\documentclass[journal]{IEEEtran}

\ifCLASSINFOpdf
\else
\fi
\usepackage{mathbbol}
\usepackage{dsfont}
\usepackage{bbm}
\usepackage{amsmath}
\usepackage{amssymb}
\usepackage{bbold}
\usepackage{mathrsfs}
\usepackage{amssymb}
\usepackage{multirow}
\usepackage{algorithm}
\usepackage{algpseudocode} 
\usepackage{graphicx}
\usepackage{amsmath}
\usepackage{booktabs}
\usepackage{graphicx}
\usepackage{enumitem}
\begin{document}

\title{Prototype-based Heterogeneous Federated
Learning for Blade Icing Detection in Wind Turbines with Class Imbalanced Data}

\author{Lele Qi, Mengna Liu, Xu Cheng,~\IEEEmembership{Senior Member,~IEEE,} Fan Shi, Xiufeng Liu, and~Shengyong Chen,~\IEEEmembership{Senior Member,~IEEE} 
\thanks{Lele Qi, Mengna Liu, Xu Cheng, Fan Shi, and Shengyong Chen are with the Engineering Research Center of Learning-Based Intelligent System (Ministry of Education), Key Laboratory of Computer Vision and System (Ministry of Education), and School of Computer Science and Engineering, Tianjin University of Technology, Tianjin, 300384, China.}
\thanks{Xu Cheng and Xiufeng Liu are with the Department of Technology, Management and Economics, Technical University of Denmark, 2800 Kongens Lyngby, Denmark.}
\thanks{Lele Qi and Mengna Liu are equal contribution. Corresponding author: Xu Cheng.}
}

\markboth{IEEE Internet of Things Journal}%
{Shell \MakeLowercase{\textit{et al.}}: Bare Demo of IEEEtran.cls for IEEE Journals}

\maketitle

\begin{abstract}
Wind farms, typically in high-latitude regions, face a high risk of blade icing. Traditional centralized training methods raise serious privacy concerns. To enhance data privacy in detecting wind turbine blade icing, traditional federated learning (FL) is employed. However, data heterogeneity, resulting from collections across wind farms in varying environmental conditions, impacts the model's optimization capabilities. Moreover, imbalances in wind turbine data lead to models that tend to favor recognizing majority classes, thus neglecting critical icing anomalies. To tackle these challenges, we propose a federated prototype learning model for class-imbalanced data in heterogeneous environments to detect wind turbine blade icing. We also propose a contrastive supervised loss function to address the class imbalance problem. Experiments on real data from 20 turbines across two wind farms show our method outperforms five FL models and five class imbalance methods, with an average improvement of 19.64\% in \( mF_{\beta} \) and 5.73\% in \( m \)BA compared to the second-best method, BiFL.
\end{abstract}


\IEEEpeerreviewmaketitle

\section{Introduction}

\IEEEPARstart{A}{n} effective strategy to reduce carbon emissions is to replace traditional fossil fuels by developing clean renewable energy sources. Among renewable energy sources, wind energy stands out as one of the most significant, alongside hydropower \cite{sadorsky2021wind}. Therefore, the efficient operation of wind turbines is crucial to maximize energy output. To optimize the harnessing of wind energy, wind farms are commonly established on ridges, mountaintops, or other elevated areas. The low-temperature climate in these areas can lead to blade icing on wind turbines, significantly reducing their efficiency, increasing maintenance costs, posing safety risks to nearby personnel and assets, and potentially causing shutdowns. In severe instances, icing on the blades can decrease global annual power generation by nearly 30\% \cite{wei2020review}. Therefore, early detection and prevention of wind turbine icing are of great significance for improving safety, power generation efficiency, and reducing maintenance costs \cite{wei2020review}.

Traditional methods for detecting ice on wind turbine blades mainly include manual observation, active de-icing technologies, and passive de-icing technologies. Manual observation relies on experiential judgment but sometimes lacks accuracy. Active de-icing technologies include heating methods \cite{wei2020review}, ultrasound \cite{overmeyer2013ultrasonic}, hot air flow \cite{wang2016light}, and electric pulses \cite{wang2020design}. these methods typically require additional energy supply, which may affect blade performance. Passive de-icing technologies involve applying coatings with special materials on the blade surface, such as black coatings, hydrophobic coatings, and chemical coatings \cite{wei2020review}. While these traditional techniques generally perform well under specific conditions, they may not fully remove ice during severe icing events. 


To overcome the limitations of traditional methods, a data-driven approach has been proposed for detecting ice on wind turbines. This method builds models by analyzing extensive data collected from supervisory control and data acquisition (SCADA) \cite{yang2013wind} systems, including operational status and environmental information.
However, traditional centralized data-driven methods can raise data privacy concerns when integrating data from various wind farms \cite{mcmahan2017communication,mothukuri2021survey,zhu2021federated}.

Traditional federated learning (FL) offers an effective solution to data privacy disclosure issue in centralized data-driven methods. Under the FL framework, each turbine contributes its own data to jointly train a global model without direct data exchange \cite{yang2019federated}. This collaborative learning method avoids centralized data storage and protects the privacy and security of data. FL has already been first applied to detect blade icing in wind turbines using a heterogeneous framework \cite{cheng2022class}.

However, there are still some limitations in applying the traditional FL framework to blade icing detection.

(1) Due to variations in geographical locations, operating environments, and hardware configurations of wind turbines, the collected data exhibit characteristics that are non-independent and identically distributed (Non-IID), thus demonstrating significant heterogeneity. This data heterogeneity severely impacts the performance of traditional FL methods in the global model for wind turbine blade icing detection. Specifically, due to the differences between turbine systems, the gradient update directions on each node are inconsistent during the training process, hindering the effective optimization of the global model during the aggregation process.

(2) Secondly, as wind turbines generally operate without issues and only infrequently experience icing, the data collected typically exhibit an imbalance. Models trained on such highly imbalanced datasets may tend to recognize the majority class, thereby overlooking critical icing anomalies. Additionally, when aggregating model parameters from different clients under an FL framework, data imbalance increases the complexity of aggregation and leads to inconsistent performance of the global model across clients, affecting the overall effectiveness and reliability of the model.

To address the above challenges, some scholars have explored exchanging encoded features, but this method still relies on server-side fusion, posing a data leakage risk \cite{cheng2022class}. We propose a federated prototype learning model for wind turbine icing detection, called FedHPb. This model communicates between clients through abstract class prototypes instead of gradients, transmitting prototypes between the server and clients without aggregating model parameters or gradients. 

The contributions of this paper are as follows:

(1) A model named FedHPb, which predicts icing on wind turbine blades using federated prototype learning, is proposed. Tailored to address the heterogeneous environments of different wind farms, FedHPb adapts effectively to various environmental conditions.  A contrastive supervised loss method is proposed to address the class imbalance problem, which does not rely on sampling the original data but instead balances the influence between classes by adjusting the dynamic weights in the loss function.

(2) The effectiveness of the FedHPb model was rigorously assessed using actual data from two distinct wind farms. Comparisons with five state-of-the-art FL models and five renowned methods for managing class imbalance unequivocally showed FedHPb's superior performance. Further, ablation experiments and sensitivity analyses were conducted to establish the critical significance of the model's principal components and parameters.

This paper is structured as follows: Section \ref{related} provides a review of the literature concerning the detection of blade icing in wind turbines and FL. Section \ref{method} details the FedHPb model that has been developed. The evaluation of the model is thoroughly discussed in Section \ref{exp}. Finally, Section \ref{conclusion} summarizes the findings and conclusions of the research.

\section{RELATED WORK}
\label{related}
\subsection{Blade Icing Detection}
Recent research on wind turbine blade icing detection mainly focuses on data-driven and traditional FL methods
Tian et al. \cite{tian2021multilevel} proposed a multi-layer convolutional recurrent neural network (MCRNN) model that combines wavelet transform with a multi-layer convolutional recurrent model. 
Tao et al. \cite{tao2023wind} proposed a new CNN-Attention-GRU model that combines the focal loss function and attention mechanism to enhance the precision of detecting icing on wind turbine blades and address data imbalance issues. 
Joyjit et al. \cite{chatterjee2023domain} proposed using a neural style transfer algorithm from generative artificial intelligence technology to  to improve the generalization capabilities of existing icing prediction models through synthetic data augmentation. 
Cheng et al. \cite{cheng2022class} pioneered the application of an FL framework featuring heterogeneous structures between clients and servers to address the blade icing detection problem, further integrating blockchain technology with FL \cite{cheng2022blockchain} to strengthen the solution. They subsequently introduced a version focused on class imbalance learning \cite{cheng2022wind}.
Zhang et al. \cite{zhang2023fedbip} proposed a FL model named FedBIP, which employs enhanced feature selection methods and segment-based oversampling techniques, and introduced an aggregation algorithm that considers each client.

Recent advances in wind turbine blade icing detection, like the methods of Tian et al. \cite{tian2021multilevel}, primarily use centralized training, which raises data privacy concerns. Although Zhang et al.'s \cite{zhang2023fedbip} traditional FL approach improves privacy protection, it inadequately addresses client data heterogeneity and imbalances in wind turbine data distribution.
To address the challenges posed by the aforementioned methods, this paper proposes a prototype-based FL framework. Compared to existing methods, this new approach not only reduces the privacy risks associated with centralized data processing but also effectively addresses data heterogeneity among clients and the imbalance in wind turbine data distribution through a prototype learning strategy tailored for imbalanced data.

\subsection{Federated Learning}
In recent years, FL has emerged as a pivotal technology for managing data privacy across multiple devices or data centers. FL is a decentralized machine learning approach that iteratively aggregates model parameters from various participants through a global server, thereby enhancing model performance and safeguarding data privacy \cite{yang2019federated}.

Despite the significant improvement in data privacy protection provided by FL, the challenge of heterogeneous data sources remains prominent. The data generated by different devices or locations often exhibit significant distribution differences, and this heterogeneity can lead to inconsistent model training outcomes, affecting overall performance. When data heterogeneity is present, directly using FedAvg  \cite{mcmahan2017communication} for parameter aggregation may result in reduced performance. In the case of model parameter heterogeneity, direct aggregation using FedAvg might be completely infeasible. To address these issues, researchers have proposed various improved aggregation methods. For instance, pFedLA \cite{ma2022layer} introduces Hypernetworks to enhance personalized learning, aiming to overcome the performance degradation of standard FL algorithms when handling client data heterogeneity. Meanwhile, FedProto \cite{tan2022fedproto} proposes a federated prototype framework that focuses on aggregating prototype representations of each class from different clients at the server side instead of model parameters. Additionally, some studies tackle FL heterogeneity by improving the objective function, such as FedProx \cite{li2020federated} which introduces regularization terms, and MOON \cite{li2021model} which employs a contrastive learning objective, further enhancing the model's adaptability and effectiveness.

\begin{figure*}[ht]
    \centering
    \includegraphics[width=1.0\linewidth]{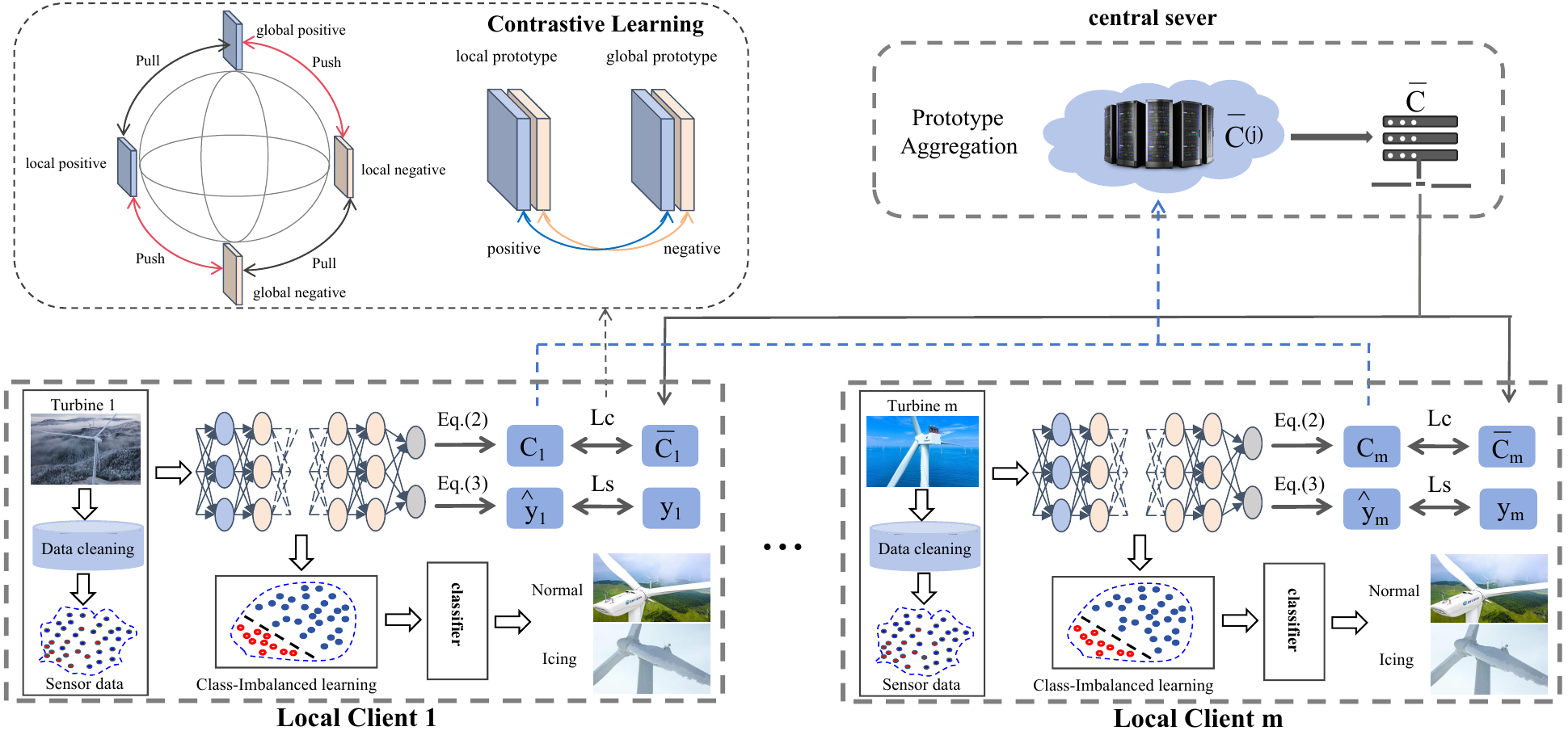}
    \caption{Overview of FedHPb: In the initial stage, clients update their local prototype sets by reducing classification errors and minimizing the differences between local and global prototypes through supervised contrastive learning. Once the updates are completed, these prototype sets are sent to the central server. The central server creates a global prototype based on the received data and sends this global prototype back to all clients.}
    \label{framework}
\end{figure*}

\begin{figure}[ht]
    \centering
    \includegraphics[width=1.0\linewidth]{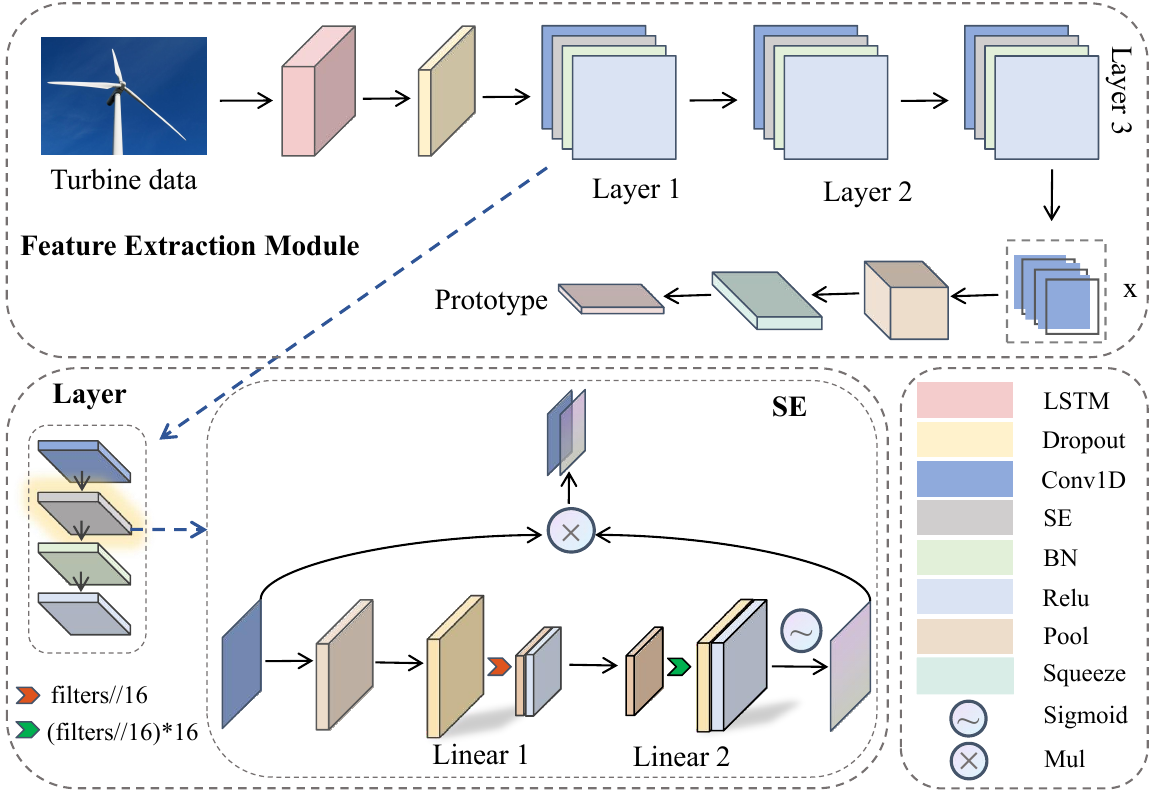}
    \caption{Neural network architecture of the client model.}
    \label{net}
\end{figure}

Recently, scholars such as Cheng et al. \cite{cheng2022class} and Zhang et al. \cite{zhang2023fedbip} have applied FL to blade icing detection in wind turbines. However, the extreme data heterogeneity among different turbine clients continues to pose significant challenges. In response, this paper introduces a prototype-based heterogeneous FL framework tailored to address this issue. This framework is specifically designed to manage the diverse data characteristics inherent to wind turbine environments.

\section{Methodology}
\label{method}
\subsection{Overview}
Wind farms are distributed worldwide, especially in mountainous and offshore areas. This widespread distribution results in the collected data exhibiting significant heterogeneity. Furthermore, icing events primarily occur in cold seasons and are relatively rare, resulting in imbalanced data that can cause models to bias towards the more frequent non-icing state. To address this, we propose an icing detection model for turbines in heterogeneous environments. Inspired by \cite{tan2022fedproto,lin2017focal,khosla2020supervised}, we propose a method for addressing class imbalance, FedHPb. As shown in Fig. \ref{framework}, the central server receives prototype sets from numerous local turbine clients. The server aggregates these prototypes by averaging them.
In the heterogeneous environment of FL, although the prototype sets from clients may overlap, they are not identical. Therefore, the server can automatically aggregate prototypes from the shared class space among clients. Additionally, the prototype-based FL strategy does not require exchanging gradients or model parameters, allowing this approach to adapt to various model architectures and effectively handle heterogeneity issues.

\subsection{Client Feature Extractor}
This article details a client model called LCNN that merges Long Short-Term Memory (LSTM) with Convolutional Neural Network (CNN). During the local training stage, each wind farm separately trains the model using its unique dataset. As shown in Fig. \ref{net}.

First, consider the original input \( X_{\text{raw}} \in \mathscr{R}^{T \times d} \), where \( d \) represents the dimension of the input and \( T \) indicates the size of the sample window. The LSTM module processes this time-series data to extract time-dependent features \cite{yin2021comprehensive}. Following the LSTM layer, a Dropout layer is implemented to boost the model's generalization capabilities and mitigate overfitting.
Subsequently, the LSTM-extracted features are fed into the CNN module, which consists of three convolutional layers. These layers integrate convolution operations, a Squeeze-and-Excitation (SE) attention mechanism, batch normalization (BN), and ReLU activation functions, collaboratively refining and enhancing the features. Finally, an adaptive average pooling layer compresses the feature map of each channel into a single value, producing a condensed, flattened feature vector. The outputs of the LCNN are represented by:
\begin{equation}
\begin{aligned}
  X_{\text{lstm}} &= \mathcal{L}(X_{\text{raw}}^{T}) \\
  X_{\text{CNN}} &= \mathcal{C}^{(3)}(X_{\text{lstm}}) \\
  X_{\text{LCNN}} &= \mathcal{P}(X_{\text{CNN}}) \\
\end{aligned}
\label{eq:full_model}
\end{equation}

Here, \( \mathcal{L}(X_{\text{raw}}^{T}) \) denotes the combined outputs of LSTM processing and the Dropout layer. The $\mathcal{C}^{(3)}$ produce the outputs from the first, second, and third convolutional layers. \( \mathcal{P} \) represents the output from the adaptive average pooling layer. The overall representation of the model is in the form of a composite function $f(X_{\text{raw}})$:
\begin{equation}
\begin{aligned}
f(X_{\text{raw}}) &= \mathcal{P} \circ \mathcal{C}^{(3)} \circ \mathcal{L}(X_{\text{raw}}^{T})
\end{aligned}
\label{full_model}
\end{equation}

The pooled result is mapped to the output class dimension through a linear layer, generating the final classification result, forming the predicted output \(\hat{y}\):
\begin{equation}
\hat{y} = \operatorname{Linear}(X_{\text{LCNN}})
\label{eq4:modified_equation}
\end{equation}

\subsection{Prototype Optimization Methods}
In this paper, we introduce a prototype method designed to tackle the challenge of data heterogeneity in detecting icing on wind turbine blades. To effectively manage the heterogeneity of different clients, we define a prototype \( C^{(j)}_i \) for each class in client i, which is the average of all instance embedding vectors belonging to class \( j \). Specifically, for each client \( i \), the prototype of class \( j \) is calculated as follows:
\begin{equation}
C^{(j)}_i = \frac{1}{\operatorname{n}^j} \sum_{k=1}^{\operatorname{n}^j} X^{i,j}_k
\label{eq5:modified_equation}
\end{equation}
where \( n^j \) is the number of instances of class \( j \) for that client, and \( X^{i,j}_k \) denotes the embedding vector of the \( k \)-th instance of class \( j \).

Furthermore, we implement a strategy that allows each local model to adjust its prototypes to align with those of other clients. This strategy aims to achieve efficient learning across heterogeneous clients by minimizing the cumulative loss across all clients in the local learning task. Let \( \mathcal{D}_i \) denote the training dataset of client \( i \), \( L_s \) be the  supervised learning loss , and \( L_c \) be the supervised contrastive loss. The optimization objective can be formulated as Equation \ref{eq:optimization}:
\begin{equation}
\begin{split}
& \operatorname{argmin}_{\theta, \{C_i\}} \sum_{i=1}^m \left[(1-\lambda) \frac{1}{|\mathcal{D}_i|} \sum_{(x, y) \in \mathcal{D}_i} L_s(\hat{y}, y) \right. \\
& \qquad + \left. \lambda \cdot \frac{1}{|\mathcal{C}|} \sum_{j=1}^{|\mathcal{C}|} L_c(C^{(j)}_i,  \overline{C}^{(j)}) \right]
\end{split}
\label{eq:optimization}
\end{equation}

where
\( C^{(j)}_i \) represents the local prototype for class \( j \) on client \( i \). \( \overline{C}^{(j)} \) denotes the global prototype for class \( j \). \( \lambda \) is the regularization coefficient that balances two losses. \( C \) signifies the total number of categories, y is the true labels.
Further details of the optimization can be found in Algorithms 1 .

\begin{algorithm}
\caption{FedHPb and LocalUpdate}
\begin{algorithmic}[1]
\State \textbf{Input:} number of communication rounds $R$, number of local epochs $E$, the number of clients $K$, temperature \( \tau \), hyper-parameter \( \lambda \)

\State \textbf{Server executes:}
\State Initialize global prototype set $\{\overline{C}^{(j)}\}$ for all classes.
\For{each round $t = 1, 2, \ldots, T$}
    \For{each client $i$ in parallel}
        \State \( C^{(j)}_i \gets \text{LocalUpdate}(i, \overline{C}^{(j)}) \)
    \EndFor
    \State Update global prototype by Eq. (\ref{juhe}).
    \State Update local prototype set $C^{(j)}_i$ with prototypes in $\{\overline{C}^{(j)}\}$
\EndFor
\end{algorithmic}

\vspace{0.5em}
\hrule
\vspace{0.5em}

\textbf{LocalUpdate} ($i, \overline{C}^{(j)})$
\begin{algorithmic}[1]
    \For{each local epoch}
        \For{batch $(x_i, y_i) \in D_i$}
            \State Compute local prototype by Eq. (\ref{eq4:modified_equation}).
            \State $L_{\text{s}} \leftarrow \text{supervised learning loss}(\hat{y}, y)$
            \State $L_{\text{c}} \leftarrow \text{supervised contrastive loss}(C^{(j)}_i,  \overline{C}^{(j)})$
            \State $L_{all} \leftarrow (1-\lambda)L_s + \lambda \cdot L_c$
            \State Update local model according to the total loss.
        \EndFor
    \EndFor
    \State \Return $C^{(j)}_i$
\end{algorithmic}
\end{algorithm}

\subsection{Local Model Update for Class-Imbalanced Data}
To guarantee high consistency of model prototypes across various clients, it is essential to synchronize the local models of each client. This study addresses this by incorporating a regularization term \( \lambda \) into the local loss function. This term is designed to minimize the discrepancy between local and global prototypes, while simultaneously reducing classification errors. The loss function comprises two primary elements: the supervised learning loss \( L_s \) and the supervised contrastive loss \( L_c \). The detailed expression for the loss function is presented as follows:
\begin{equation}
L_{all} = (1-\lambda)L_s + \lambda \cdot L_c
\label{eq6}
\end{equation}

The supervised learning loss \( L_s \) is computed in the following manner:
\begin{equation}
L_s(\hat{y}, y) = - \sum_{i=1}^{N} \sum_{j=1}^{C} \left( \frac{N}{C \cdot {n}^j} \right) \cdot y_{i,j} \log(\hat{y}_{i,j})
\end{equation}
In this formula, \( N \) denotes the total number of samples, \( y_{i,j} \) represents the true label indicating that the \( i \)-th sample belongs to class \( j \). This strategy helps to balance the loss function, ensuring that the model gives greater emphasis to minority classes.

Inspired by contrastive learning, the proposed supervised contrastive loss \( L_c \) aims to enhance the model's ability to differentiate between classes, particularly when dealing with class imbalance in the icing dataset. By bringing similar samples closer and pushing dissimilar samples apart, this loss function optimizes the model. \( L_c \) is defined as follows:
\begin{equation}
L_c (C^{(j)}_i,  \overline{C}^{(j)}) = \left( w_{\text{pos}} \cdot {L_{\text{pos}}} + w_{\text{neg}} \cdot {L_{\text{neg}}} \right)
\end{equation}

By assigning higher weights to the minority icing samples, the model's capability to identify minority classes is improved. The positive and negative sample losses \( L_{\text{pos}} \) and \( L_{\text{neg}} \) are defined as follows:
\begin{equation}
\small
\begin{aligned}
L_{\text{pos}} &= -\log \left( \frac{e^{\operatorname{sim}(C^{(0)}_i, \overline{C}^{(0)}) / \tau}}{e^{\operatorname{sim}(C^{(0)}_i, \overline{C}^{(0)}) / \tau} + e^{\operatorname{sim}(C^{(0)}_i, \overline{C}^{(1)}) / \tau} + \epsilon} \right) \\
L_{\text{neg}} &= -\log \left( \frac{e^{\operatorname{sim}(C^{(1)}_i, \overline{C}^{(1)}) / \tau}}{e^{\operatorname{sim}(C^{(1)}_i, \overline{C}^{(1)}) / \tau} + e^{\operatorname{sim}(C^{(1)}_i, \overline{C}^{(0)}) / \tau} + \epsilon} \right)
\end{aligned}
\label{eq1:loss_functions}
\end{equation}

Here, \( C^{(0)}_i \) and \( C^{(1)}_i \) represent the feature vectors for local positive and negative samples, respectively. Similarly, \( \overline{C}^{(0)} \) and \( \overline{C}^{(1)} \) denote the feature vectors for global positive and negative samples. The subscript \( j \) in \( C^{(j)}_i \) and \( \overline{C}^{(j)} \) indicates the categories: non-icing when \( j \) is 0 and icing when \( j \) is 1.
The similarity function \( \text{sim} \) computes the similarity between two vectors. The temperature parameter \( \tau \) scales the similarity scores. To prevent division by zero, the term \( \epsilon \) is added to the denominator.


The weights \( w_{\text{pos}} \) and \( w_{\text{neg}} \) are based on the number of instances per class label, adjusting the loss function's impact. Specifically, \( n^0 \) represents the count of non-icing labels, and \( n^1 \) represents the count of icing labels, \( \epsilon \) is a very small value introduced to prevent division by zero errors. \( \gamma \) is a parameter that adjusts the sensitivity of the weights, and it is set to 2 in this context:
\begin{equation}
w_{\text{pos}} = \left( \frac{1}{n^0 + \epsilon} \right)^{\gamma} \quad w_{\text{neg}} = \left( \frac{1}{n^1 + \epsilon} \right)^{\gamma}
\end{equation}

This method improves model classification by integrating global and local prototype updates, contrastive learning loss, and dynamic weight adjustment. The first supervised loss boosts accuracy and fairness by adjusting minority class weights, while the second supervised contrastive loss enhances intra-class similarity and inter-class differences, improving class differentiation.

\subsection{Global Prototype Aggregation}
Given the diversity in data and models among clients, each requires tailored model parameters for optimal performance, a need inadequately served by standard gradient-based communication. However, the commonality of label spaces across clients allows them to function within a unified embedding space. This facilitates efficient information exchange among heterogeneous clients through the strategic aggregation of class-specific prototypes.

For class \( j \), the server collects prototypes from clients that hold data for class \( j \) and constructs a global prototype, \(\overline{C}^{(j)}\), for that class through a prototype aggregation process. This process is performed by calculating the weighted average of each label's global prototype, with the weights determined by the size of each client's label dataset. The precise formula is as follows:
\begin{equation}
\label{juhe}
\overline{C}^{(j)} = \frac{1}{|N_j|} \sum_{i \in N_j} \frac{|D_{i,j}|}{|N_j|} C_{i}^{(j)}
\end{equation}

where \(C_{i}^{(j)}\) is the local prototype of label \( j \) for client \( i \). \(|N_j|\) is the number of clients that have label \( j \). \(D_{i,j}\) is the size of the label \( j \) dataset for client \( i \). Through this method, the server can effectively share and exchange information among heterogeneous clients, improving the overall performance and generalization capability of the model.

\section{EXPERIMENT}
\label{exp}
The models in this study were implemented using Pytorch (version 1.9.0). All experiments were performed on a server equipped with an NVIDIA GeForce RTX 4090 GPU. The following hyperparameter settings were used during model training:

(1) The Stochastic Gradient Descent (SGD) optimizer was utilized with a learning rate set at 0.01.

(2) The hyper-parameter \( \lambda \) was set to 0.25, and the temperature \( \tau \) was set to 0.5.

(3)The communication rounds (\( R \)) between clients and the server were configured to 20, while the local training epochs (\( E \)) were determined to be 100.

\subsection{Experimental Setup and Data}
Data for this study were obtained from two wind farms\cite{cheng2022class}, which are located approximately 700 kilometers apart, specifically in Hongshimu, Shaanxi Province, and Yanling, Henan Province. A total of 20 wind turbines were selected for data collection, with 10 turbines from each wind farm. The data from the Hongshimu wind farm were collected from February 12, 2019, to February 26, 2019, spanning 15 days, with one day containing icing data. The data from the Yanling wind farm were collected from February 11, 2019, to February 26, 2019, spanning 16 days, with two days containing icing data. Wind turbine experts identified 16 variables related to blade icing, and the data were labeled as either icing or non-icing. 

Prior to data analysis, preprocessing steps such as denoising, outlier handling, and missing value imputation were performed to ensure data quality and consistency. The data from both wind farms are representative and reflect local climatic parameters, making them suitable for training a wind turbine icing detection network model that can adapt to different climatic conditions. Consequently, all experiments in this paper are based on this dataset.

In terms of experimental design, the data were divided into a training set (60\%) and a testing set (40\%). The training set was utilized to construct the model, whereas the testing set was employed to assess the model's performance. To comprehensively assess the model's adaptability to different levels of data imbalance, we explored three imbalance ratios (i.e., the ratio of non-icing samples to icing samples): 20:1, 50:1, and 100:1. Given the sporadic nature of icing events, we specifically set a sample imbalance ratio of 10:1 in the testing set. This setup aims to test the model's performance and detection accuracy when facing imbalanced data.

\subsection{Evaluation Indicators}
In this study, for the binary classification task of wind turbine icing prediction, we selected the \( F_{\beta} \) score and balanced accuracy (BA) as the main performance evaluation metrics. In this experiment, the value of \( \beta \) is set to 2. The \( F_{\beta} \) score is calculated as follows:
\begin{equation}
F_\beta = \frac{(1 + \beta^2) \cdot \text{Precision} \cdot \text{Recall}}{\beta^2 \cdot \text{Precision} + \text{Recall}}
\end{equation}

Here, Precision is calculated as the ratio of correctly predicted icing events to all events predicted as icing, Recall is determined as the ratio of correctly predicted icing events to all actual icing events. The formulas are as follows:
\begin{equation}
\text{Precision} = \frac{\text{TP}}{\text{TP} + \text{FP}} \quad \text{Recall} = \frac{\text{TP}}{\text{TP} + \text{FN}}
\end{equation}

Balanced accuracy (BA) is used as a metric to evaluate the model's performance in imbalanced data scenarios, equally considering the model's ability to recognize positive (non-icing) and negative (icing) classes. It is defined as follows:
\begin{equation}
\text{BA} = \frac{1}{2} \left( \frac{\text{TP}}{\text{TP} + \text{FN}} + \frac{\text{TN}}{\text{TN} + \text{FP}} \right)
\end{equation}

Here, TP, TN, FP, and FN denote true positives, true negatives, false positives, and false negatives, respectively.

For \( K \) client models, we calculated the average \( F_{\beta} \) score and balanced accuracy (BA). The global average \( F_{\beta} \) score ( \( mF_{\beta} \) ) is the arithmetic mean of all client \( F_{\beta} \) scores, and the global balanced accuracy ($m$BA) is the arithmetic mean of all client balanced accuracies:
\begin{equation}
mF_{\beta} = \frac{1}{K} \sum_{i=1}^{K} F_{\beta}^{i}  \quad m\text{BA} = \frac{1}{K} \sum_{i=1}^{K} \text{BA}^{i}
\end{equation}

To mitigate the effect of randomness, we computed the average value across all server training rounds within each communication round.

\subsection{Comparison With State-of-The-Art FL Methods}
In this study, we introduced and evaluated five different FL frameworks.
\begin{enumerate}[label=\arabic*)] %
  \item \textbf{FedAvg} (Federated Mean) \cite{mcmahan2017communication}: In the FedAvg algorithm, clients train models on local data and upload the model parameters to a server, which then aggregates these parameters using a weighted average method.
  \item \textbf{FedProto} (Federated Prototype) \cite{tan2022fedproto}: The FedProto framework improves tolerance to heterogeneity by transmitting abstract class prototypes between clients and servers, instead of gradients.
  \item \textbf{FedBN} (FL on Non-IID Features Via Local Batch Normalization) \cite{li2021fedbn}: FedBN tackles Non-IID data issues using a local batch normalization strategy. Nodes retain BN layer parameters and upload others for global aggregation.
   \item \textbf{FLGKT} (FL with Global Knowledge Transfer) \cite{he2020group}: This framework improves model performance on heterogeneous and non-independent identically distributed data by promoting knowledge transfer among clients.
   \item \textbf{BIFL} (Blade Icing FL) \cite{cheng2022class}: This method proposes a heterogeneous wind turbine icing detection model between client and server.
\end{enumerate}


In this experiment, a sampling window size of 128 was used for data collection. As shown in Table \ref{tab:my_label}, the results indicate that FedHPb outperformed all other methods across the three imbalance ratios, particularly excelling in the \( mF_{\beta} \) metric. This suggests that FedHPb is effective in addressing class imbalance issues. In contrast, FLGKT and BiFL also demonstrated better performance, particularly at lower imbalance ratios. However, FedAvg and FedProto performed poorly in handling class imbalance, with performance significantly deteriorating at higher imbalance ratios. Our model notably outperformed the second-best method, BiFL, achieving average improvements of 19.64\% and 5.73\% in the \( mF_{\beta} \) and $m$BA metrics, respectively, across all three imbalance ratios. Even at the highest imbalance ratio of 100:1, where all methods experienced a decline in performance, our model maintained high \( mF_{\beta} \) and $m$BA values (73.08\% and 85.35\%, respectively). This demonstrates that the accuracy of all methods decreases as the imbalance ratio increases. A key factor contributing to the superior performance of our model is that BiFL uses a knowledge preservation mechanism that adds complexity. However, if the encoding of the prototype feature maps is not sufficiently refined, this may degrade prototype quality, introduce bias, and negatively affect performance.
\begin{table}[h]
\centering
\caption{PERFORMANCE COMPARISON WITH FL METHODS.}
\label{tab:my_label}
\begin{tabular}{lccc}
\toprule
Method & $\rho = 20:1$ & $\rho = 50:1$ & $\rho = 100:1$ \\
\midrule
       & $mF_{\beta}$ \ \ $m$BA & $mF_{\beta}$ \ \ $m$BA & $mF_{\beta}$ \ \ $m$BA \\
\midrule
FedAvg & 45.56 \ \ 70.01 & 38.64 \ \ 64.37 & 37.95 \ \ 63.74 \\
FedProto   & 36.61  \ \ 63.70 & 33.27 \ \ 59.68 & 33.17 \ \ 58.50  \\
FedBN   &78.21   \ \ 87.50  & 52.11 \ \ 73.58 &35.27 \ \ 65.40  \\
FLGKT  & 80.14 \ \ 91.05 & 61.28 \ \ 80.70 & 58.02 \ \ 76.43 \\
BiFL   & 77.72 \ \ 90.25 & 65.76 \ \ 84.77 & 64.76 \ \ 83.82 \\
Ours   & \textbf{91.61} \ \ \textbf{96.53} & \textbf{84.30} \ \ \textbf{91.88} & \textbf{73.08} \ \ \textbf{85.35} \\
\bottomrule
\end{tabular}
\end{table}
\subsection{Comparison With State-of-The-Art Class Imbalance Methods}
We compared the proposed FedHPb with the following five class-imbalanced algorithms.

\begin{enumerate}[label=\arabic*)]
  \item \textbf{Focal} (Focal Loss): This is a commonly used loss function in class imbalance learning \cite{lin2017focal}. We set \( \gamma = 1 \) and applied uniform weights across all classes.
  \item \textbf{CB} (Class Balance): This method introduces the concept of effective sample number, considering data overlap. It calculates CB loss based on the effective sample number for each class \cite{cui2019class}.
  \item \textbf{GHMC} (Gradient Harmonizing Mechanism Classification): This method is introduced to tackle data imbalance issues and address inconsistencies in imbalanced classification \cite{li2019gradient}.
  \item \textbf{LDAM} (Label-Distribution-Aware Margin): This method can substitute  the standard cross-entropy objective and complement traditional class imbalance strategies like re-weighting and re-sampling \cite{cao2019learning}.
  \item \textbf{Balance}: This method uses a data-level algorithm to perform undersampling on the majority class \cite{leevy2018survey}.
\end{enumerate}

As shown in \ref{tab:my_label2}, our model excels in both the m\( F_{\beta} \) and $m$BA metrics across three imbalance ratios, outperforming algorithms such as Focal, CB, GHMC, LDAM, and Balance. The Balance method excels with lower class imbalances but its performance significantly declines as the imbalance rate increases. Compared to the less optimal Balance algorithm, our model shows a significant improvement in average performance on datasets with three imbalanced ratios, boosting the m\( F_{\beta} \) and $m$BA metrics by 24.27\% and 6.78\%, respectively. This demonstrates its exceptional effectiveness in managing varying degrees of data imbalance. The Balance algorithm underperforms in wind turbine icing detection primarily because its undersampling of the majority class, which can lead to significant information loss, especially when icing events are infrequent. 
\begin{table}[h]
\centering
\caption{Performance Comparsion With State-Of-The-Art Class-Imbalanced Methods}
\label{tab:my_label2}
\begin{tabular}{lcccccc}
\toprule
Method & \multicolumn{2}{c}{$\rho = 20:1$} & \multicolumn{2}{c}{$\rho = 50:1$} & \multicolumn{2}{c}{$\rho = 100:1$} \\
\cmidrule(r){2-3} \cmidrule(lr){4-5} \cmidrule(l){6-7}
       & $mF_{\beta}$ & $m$BA & $mF_{\beta}$ & $m$BA & $mF_{\beta}$ & $m$BA \\
\midrule
Focal  & 25.74 & 55.38 & 21.68 & 53.18 & 24.82 & 54.41 \\
CB     & 44.55 & 68.88 & 43.91 & 68.37 & 43.09 & 67.42 \\
GHMC   & 27.94 & 51.21 & 26.92 & 50.12 & 27.11 & 50.27 \\
LDAM   & 18.96 & 52.35 & 16.73 & 51.38 & 16.31 & 50.86 \\
Balance& 72.15 & 88.83 & 65.66 & 85.04 & 62.22 & 82.36 \\
Ours   & \textbf{91.61} & \textbf{96.53} & \textbf{84.30} & \textbf{91.88} & \textbf{73.08} & \textbf{85.35} \\
\bottomrule
\end{tabular}
\end{table}

\subsection{Comparison of Communication Efficiency}
In this study, we compare the network efficiency of two FL models during the training process.
The methodology employed entails quantifying the average size of serialized model updates transmitted by all clients in each communication round, which provides insights into the network resource consumption of each model. 


\begin{figure}[h]
    \centering
    \includegraphics[width=0.65\linewidth]{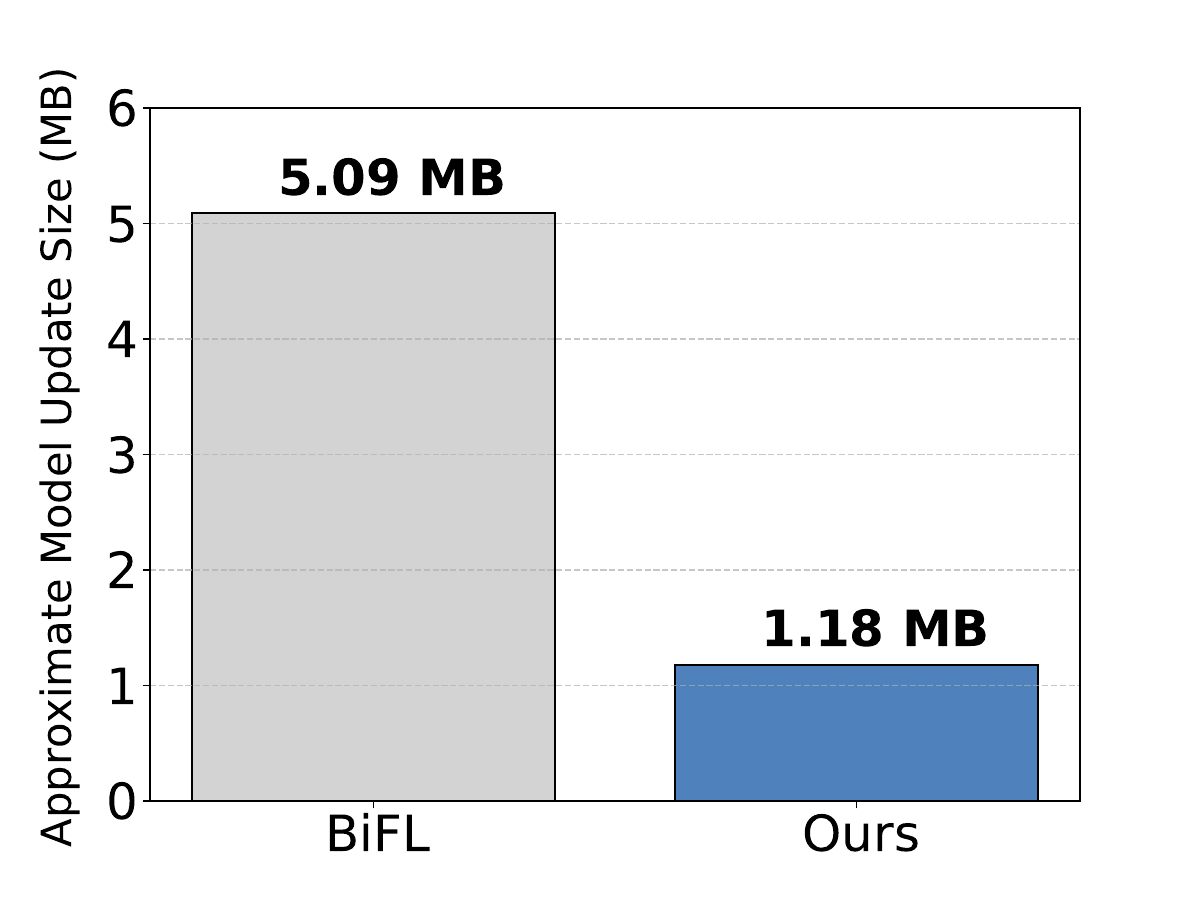}
    \caption{Approximate Model Update Size (MB)}
    \label{tab:sparsity_performance}
\end{figure}

As shown in Fig. \ref{tab:sparsity_performance}, the average data transmitted per client per communication round by our model is approximately 1.18 MB, significantly lower than the 5.09 MB recorded for the BiFL model. This substantial reduction in per-client data transmission, while preserving robust performance, underscores the enhanced efficiency of our model in FL scenarios.

\subsection{Ablation and Sensitivity Analysis}
\subsubsection{\textbf{Influence of the supervised contrastive loss \( L_c \)}}
To address the issue of data imbalance in wind turbine icing detection, we introduce a supervised contrastive loss function. Moreover, a practical approach involves integrating regularization into the local loss component. In this context, the \( L_2 \) distance denotes the Euclidean distance between the local prototype \( C^{(j)}_i \) and the global prototype \( \overline{C}^{(j)} \). The \( L_2 \) regularization is specified as follows:
\begin{equation} 
L_2 = \left\|  C^{(j)}_i  - \overline{C}^{(j)} \right\|_2 
\end{equation}

We compared different types of loss functions to constrain the representations: without additional terms (\( L_{all}= L_{\text{s}} \)), \( L_2 \) norm (\( L_{all} = (1-\lambda)L_{\text{s}} + \lambda L_2 \)), and our model loss (\( L_{all} = (1-\lambda)L_{\text{s}} + \lambda L_c \)).

The results, as shown in Table \ref{tab:my_label3}, indicate that introducing the \( L_2 \) norm significantly reduces performance across different data imbalance ratios compared to using no normalization. While \( L_2 \) regularization promotes smoothness in the learned representations, it does not prioritize class boundaries. When applied to imbalanced datasets, it causes the model to focus on the dominant class, resulting in suboptimal performance for the minority class. In contrast, our method enhances \( mF_{\beta} \) and \( m \)BA, demonstrating its effectiveness in handling class-imbalanced data.

\begin{table}[h]
\centering
\caption{Performance Comparison of The supervised contrastive loss \( L_c \)}
\label{tab:my_label3}
\begin{tabular}{lccc}
\toprule
second term & $\rho = 20:1$ & $\rho = 50:1$ & $\rho = 100:1$ \\
\midrule
       & $mF_{\beta}$ \ \ $m$BA & $mF_{\beta}$ \ \ $m$BA & $mF_{\beta}$ \ \ $m$BA \\
\midrule
none & 91.59 \ \ 96.44 & 83.07 \ \ 91.17 & 73.04 \ \ 85.12 \\
\( L_2 \) norm & 86.96 \ \ 94.76 & 78.91 \ \ 89.92 & 70.11 \ \ 84.28 \\
Ours   & \textbf{91.61} \ \ \textbf{96.53} & \textbf{84.30} \ \ \textbf{91.88} & \textbf{73.08} \ \ \textbf{85.35} \\
\bottomrule
\end{tabular}
\end{table}

\subsubsection{\textbf{Influence of the Modules in Local Model}}
To evaluate the effectiveness of the various components in the proposed model, this section compares the FedHPb model with its modified variants. The following variants have been derived from the FedHPb model:
\begin{enumerate}[label=\arabic*)]
\item  FedHPb\_SiLU: Replaces the activation function in the CNN part of the model with the SiLU function.
\item  FedHPb\_ADAM: Replaces the optimizer in the model with ADAM.
\end{enumerate}

\begin{figure}[h]
    \centering
    \begin{minipage}{0.45\linewidth}
        \centering
        \includegraphics[width=\linewidth]{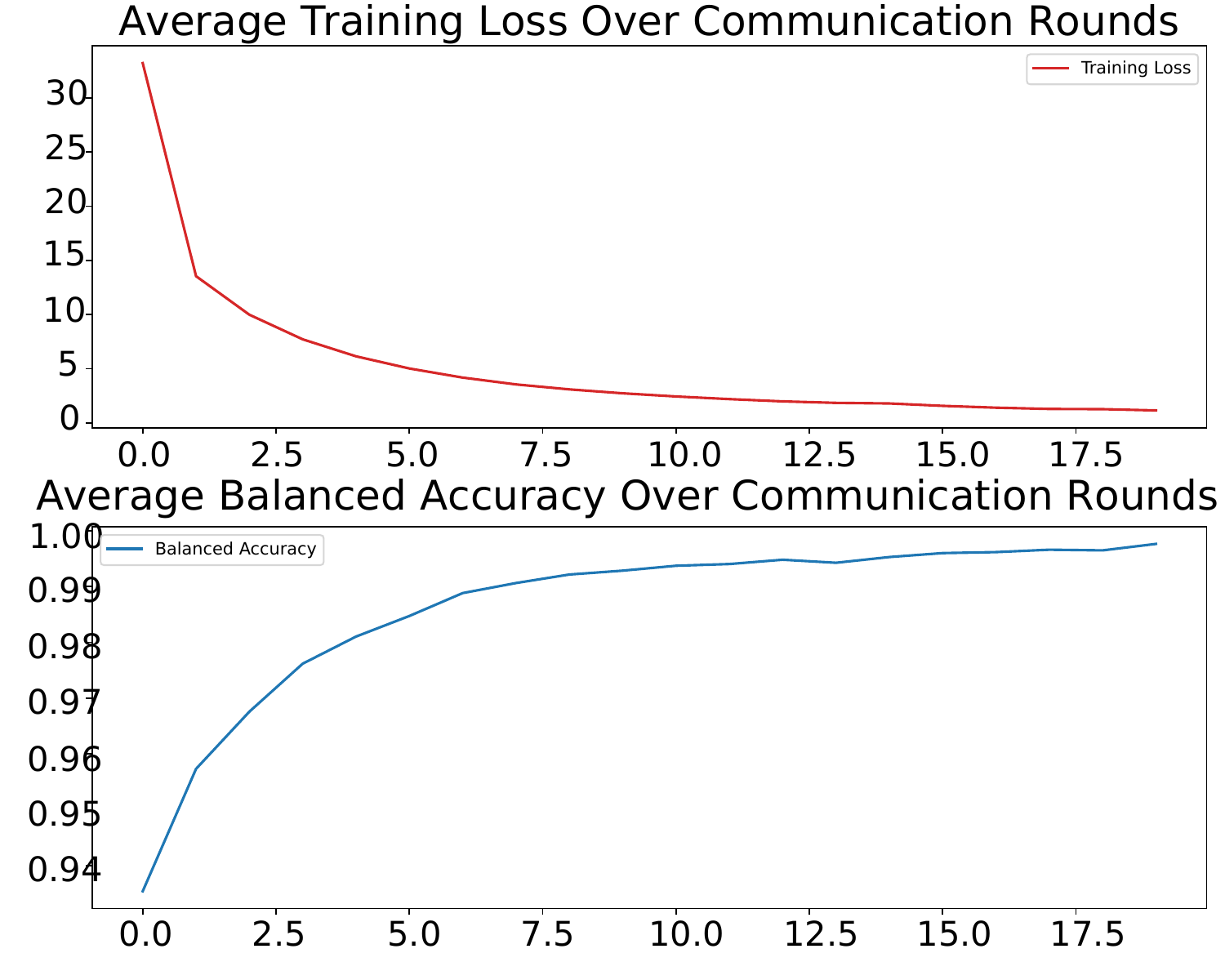}
        \caption{SGD variation curve.}
        \label{SGD}
    \end{minipage}
    \hfill
    \begin{minipage}{0.45\linewidth}
        \centering
        \includegraphics[width=\linewidth]{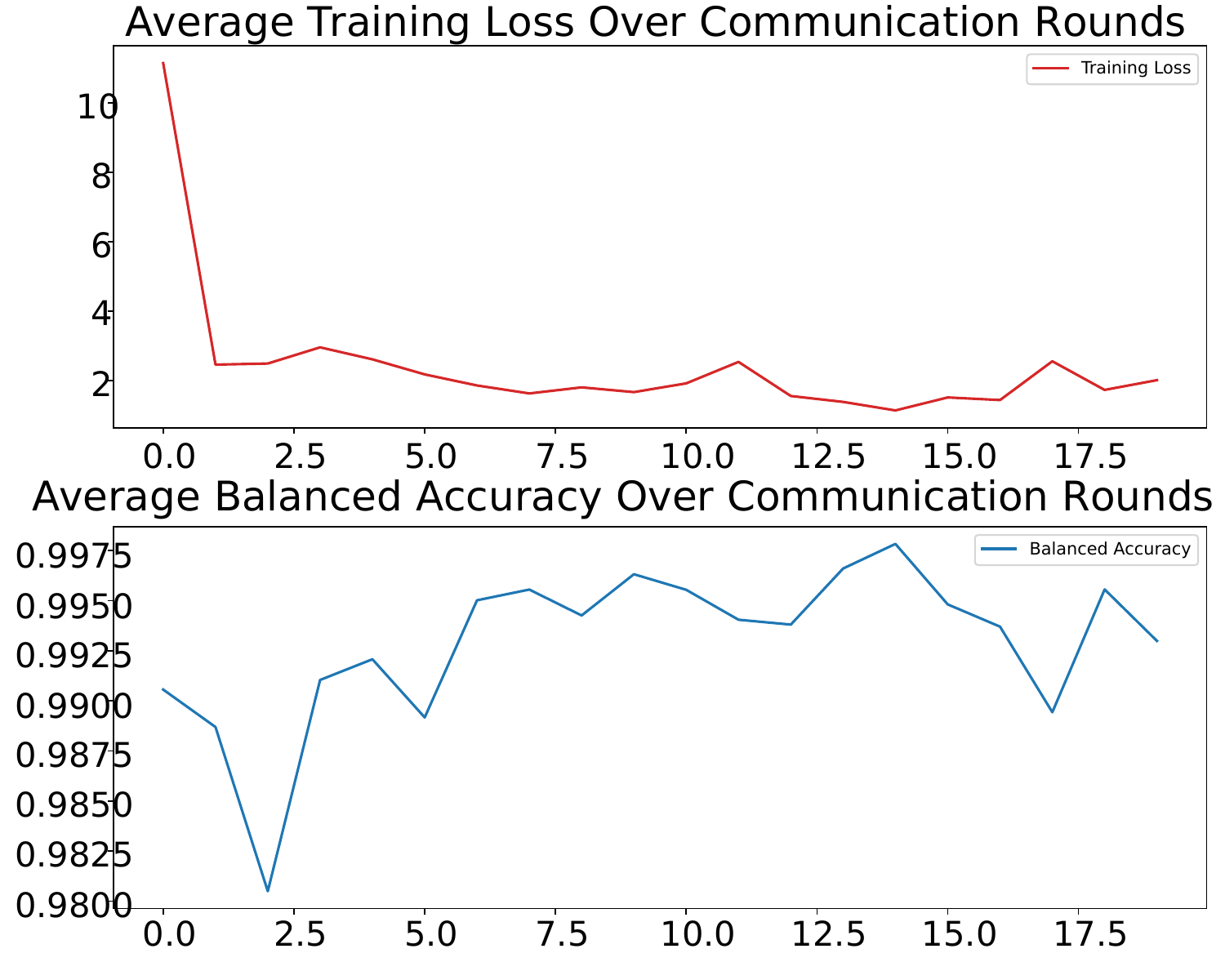}
        \caption{ADAM variation curve.}
        \label{ADAM}
    \end{minipage}
\end{figure}

Based on the experimental results in Table \ref{tab:my_label4}, we analyze the performance of different model variants. The results indicate that, compared to using ReLU as the activation function, the model with the SiLU activation function exhibits an average decrease of 2.49\% in \( mF_{\beta} \) and 0.91\% in \( m \)BA across three imbalance ratios. ReLU activation functions may mitigate gradient vanishing and enhance training efficiency in shallow networks, whereas SiLU activation functions, which promote smoother gradient flow, are better suited for deeper networks. Additionally, compared to using SGD as the optimizer, the model with the ADAM optimizer shows an average decrease of 6.22\% in \( mF_{\beta} \) and 3.63\% in \( m \)BA across the three imbalance ratios. Therefore, the proposed modules and their combinations are effective.
As shown in Fig. \ref{SGD} and Fig. \ref{ADAM}, the model using the SGD optimizer exhibits a smoother decrease in loss and a steady increase in balanced accuracy (BA) under the same parameter settings. In contrast, the model using the ADAM optimizer shows larger fluctuations in loss and more unstable BA performance during training. In the context of FL for wind turbine icing detection, if each client in a distributed training process experiences fluctuations while using Adam, it could slow down the training speed and reduce the accuracy of the global model.
\begin{table}[h]
\centering
\caption{Ablation Study}
\label{tab:my_label4}
\begin{tabular}{lccc}
\toprule
Method & $\rho = 20:1$ & $\rho = 50:1$ & $\rho = 100:1$ \\
\midrule
       & $mF_{\beta}$ \ \ $m$BA & $mF_{\beta}$ \ \ $m$BA & $mF_{\beta}$ \ \ $m$BA \\
\midrule
FedHPb\_SiLU & 89.65 \ \ 95.85 & 81.04 \ \ 90.44 & 72.17 \ \ 84.98 \\
FedHPb\_ADAM & 87.83 \ \ 93.63 & 79.23 \ \ 88.56 & 67.70 \ \ 82.04 \\
Ours   & \textbf{91.61} \ \ \textbf{96.53} & \textbf{84.30} \ \ \textbf{91.88} & \textbf{73.08} \ \ \textbf{85.35} \\
\bottomrule
\end{tabular}
\end{table}

\subsubsection{\textbf{Sensitivity Analysis of Window Size}}
\begin{figure}[h]
    \centering
    \begin{minipage}{0.333\linewidth}
        \centering
        \includegraphics[width=\linewidth]{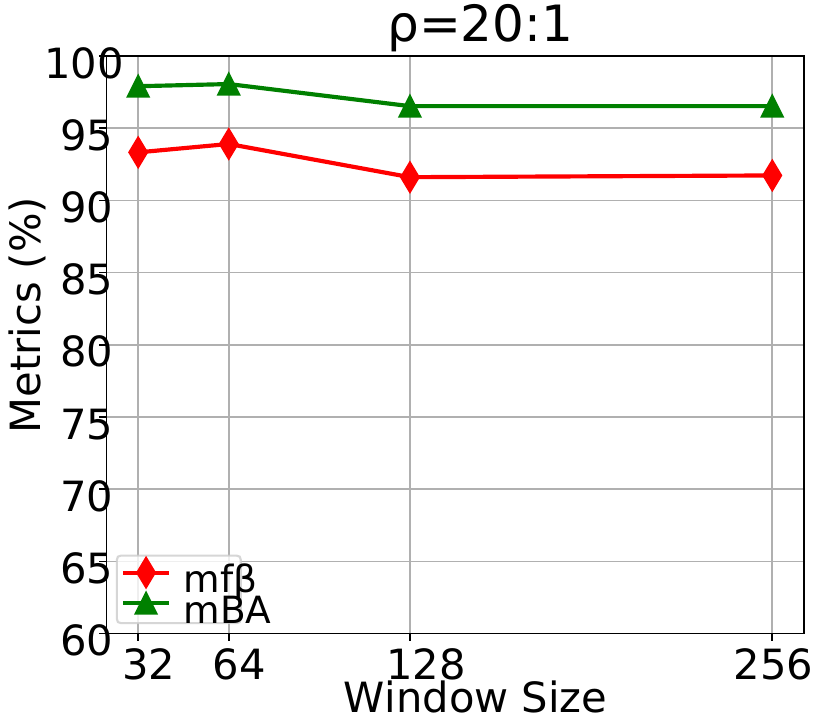}
    \end{minipage}%
    \begin{minipage}{0.333\linewidth}
        \centering
        \includegraphics[width=\linewidth]{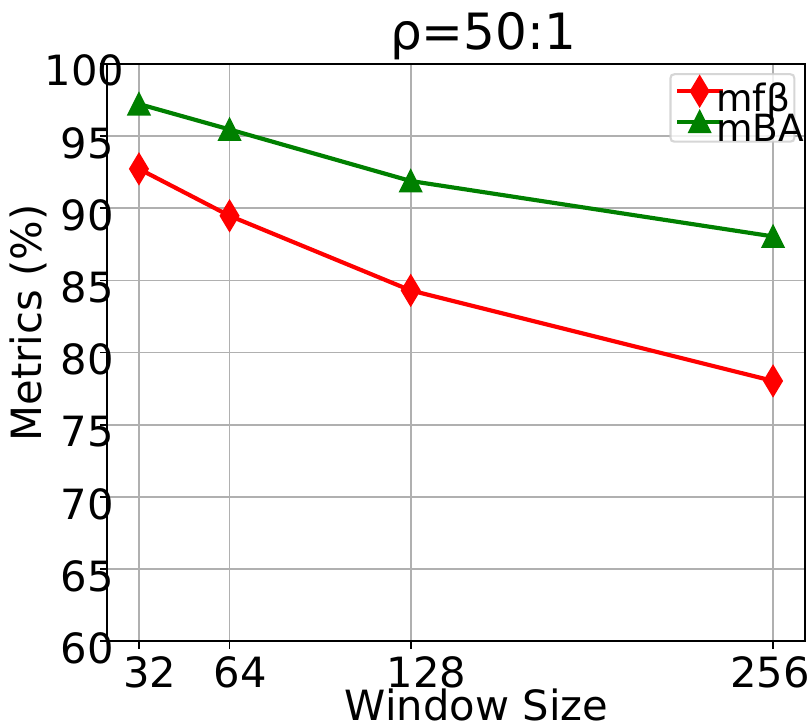}
    \end{minipage}%
    \begin{minipage}{0.333\linewidth}
        \centering
        \includegraphics[width=\linewidth]{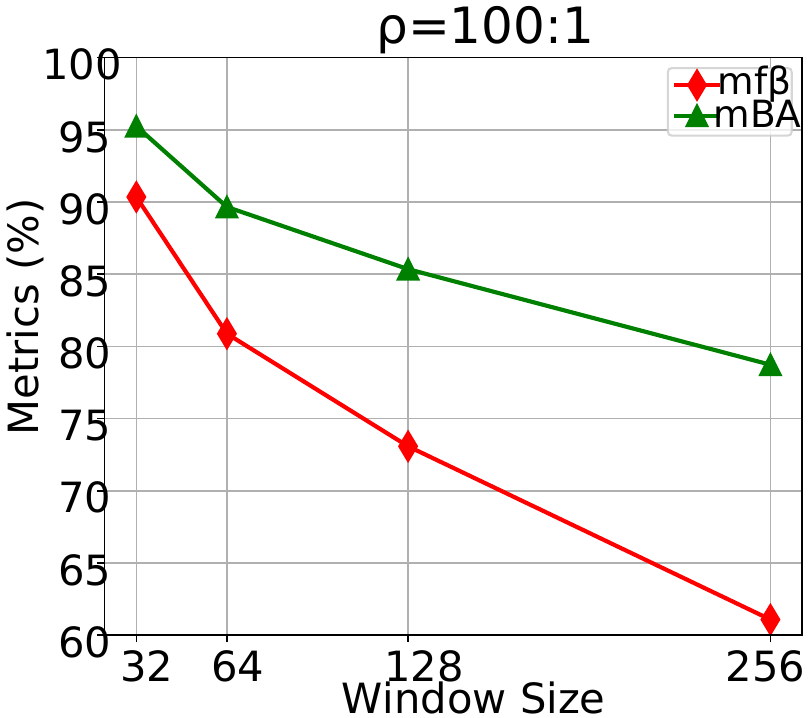}
    \end{minipage}
    \caption{Sensitivity analysis of window size.}
    \label{sen}
\end{figure}

To assess the impact of input window size, we performed a sensitivity analysis by varying the window size from 32 to 256. As shown in Fig. \ref{sen}, the performance metrics \( mF_{\beta} \) and $m$BA are optimal at smaller window sizes (32). As the window size increases to 256, the performance generally declines. In scenarios where there is an imbalance ratio of 20:1, these two metrics show a slight decrease with increasing window size but remain relatively stable overall. However, as the imbalance ratio increases to 50:1 and 100:1, the performance drop becomes more significant, especially at the maximum window size (256). This suggests that smaller window sizes are better suited for tasks with high data imbalance, as they help the model focus on more relevant, short-term patterns and reduce the impact of irrelevant long-term dependencies.

\section{Conclusion}
\label{conclusion}
This paper introduces FedHPb, a novel class-imbalanced heterogeneous FL model designed for detecting icing issues on wind turbine blades. Considering that blade icing data is commercially sensitive information, FedHPb employs FL technology to ensure data privacy is protected. Additionally, to tackle the class imbalance in the training data, the study introduces a prototype-based method combined with contrastive learning techniques. A comprehensive evaluation of the proposed model was conducted by comparing it with five traditional FL models and five advanced class imbalance learning methods. Experimental results demonstrate the model's substantial superiority. Moreover, ablation studies confirmed the effectiveness of the FedHPb modules, while sensitivity analyses explored the influence of key hyperparameters.

Future work will focus on several directions. We are dedicated to minimizing data exposure risk, particularly when clients send extracted features to the server for global prototype aggregation. Confusion matrix techniques and homomorphic encryption are effective for protecting data privacy.

\bibliographystyle{IEEEtran}
\bibliography{IEEEabrv, ./bib/paper}

\end{document}